# An Adaptive Sampling Scheme to Efficiently Train Fully Convolutional Networks for Semantic Segmentation

Lorenz Berger[1,2], Eoin Hyde[1,2], M. Jorge Cardoso[1], and Sébastien Ourselin[1]

[1] TIG, CMIC, University College London, London, UK
[2] Innersight Labs, London, UK

## Abstract

Deep convolutional neural networks (CNNs) have shown excellent performance in object recognition tasks and dense classification problems such as semantic segmentation. However, training deep neural networks on large and sparse datasets is still challenging and can require large amounts of computation and memory. In this work, we address the task of performing semantic segmentation on large data sets, such as three-dimensional medical images. We propose an adaptive sampling scheme that uses a-posterior error maps, generated throughout training, to focus sampling on difficult regions, resulting in improved learning. Our contribution is threefold: 1) We give a detailed description of the proposed sampling algorithm to speed up and improve learning performance on large images. 2) We propose a deep dual path CNN that captures information at fine and coarse scales, resulting in a network with a large field of view and high resolution outputs. 3) We show that our method is able to attain new state-of-the-art results on the VISCERAL Anatomy benchmark.

## 1 Introduction

This paper addresses the problem of efficiently training convolutional neural networks (CNNs) on large and imbalanced datasets. We propose a training strategy that adaptively samples the training data to effectively speed up training and avoid over-sampling data that contains little extra information.

In this work, we investigate the problem of automatic segmentation from high resolution 3D CT scans. Several deep learning techniques [1–4] have recently been proposed for 3D segmentation of medical datasets. To overcome the problem of dealing with these large datasets, such as Computed Tomography (CT) volumes, commonly of dimension $512 \times 512 \times 700$, previous approaches train a CNN on a cropped region of interest which reduces the size of individual training images by around 100 fold [2, 4]. By reducing the size of training images, they can now be fit into memory and a network can be trained effectively on the selected data. However, identifying regions of interest requires an additional pre-processing step which may not be easy in many applications. Also, training CNNs on cropped images limits the field of view of the CNN and subsequently can introduce unwanted image boundary induced effects during testing. Other applications, where training CNNs on very large images is a problem, includes the segmentation of histology datasets [5] or the segmentation of aerial images. For example in aerial image segmentation, training a CNN to segment ships [6] can be difficult because large portions of the image contain water which provide little information during training, resulting in slow learning. Some ideas to address this have already been proposed, for example in [3] a fixed, hand-crafted, pre-computed weight map is used to help learn small separation borders between touching cells for biomedical image segmentation. In this work the proposed sampling scheme ends up dynamically learning such a weight mapping, making it generally applicable to many applications.

Curriculum learning [7] and derivative methods like self-paced learning [8] build on the intuition that, rather than considering all samples simultaneously, the algorithm should be presented with the training data in a meaningful order that facilitates learning. These ideas have already successfully been applied to image classification [9, 10], by ordering images from easy to hard during training. Also for the problem of weakly supervised semantic segmentation [11] similar ideas are applied, where predictions from previous training iterations are used to iteratively learn segmentation maps from just a single class label per image.

The focus of this paper is fully supervised semantic segmentation where a representative training set is available with dense manual label annotations and the challenge lies in efficiently learning from this large datasets. We give a detailed account of the implementation, which is a straightforward extension to any existing CNN segmentation system, and present state-of-the-art segmentation results on the VISCERAL anatomy benchmark.

## 2 Methods

### 2.1 Neural Network Architecture

For the dual path network architecture we build on several previous ideas [1, 12, 13]. Compared to the 3D network outlined in [1], we further develop the architecture by replacing the standard convolution layers with popular resnet blocks [14], and increase the maximum network depth from 11 layers in [1] to 19 layers. By having a deeper network and a down sampled pathway with input resolution 1/4 of the original resolution, we obtain a large receptive field of size $124^3$ whilst maintaining a deep high resolution pathway that does not compromise the resolution through pooling layers. The architecture results in a total of 649,251 parameters. A sketch of the architecture is given in Figure 1a. In Figure 1a, numbers inside round brackets give the input dimensions of each block. For the training stage these dimensions have been chosen carefully to balance memory usage with processing speed. During testing the dimensions may be chosen as large as can be fit into memory, to take advantage of the fully convolutional inference. Numbers in square brackets refer to the number of feature maps used at each layer. The proposed configuration allows for a large number of samples ($3D$ patches) per batch to ensure balanced class sampling and effective optimization, whilst maintaining a deep and wide enough network to capture the high variability and spatial semantics of the data. The blocks labeled 'Conv' are standard convolutional layers with kernel size $3 \times 3 \times 3$. The blocks labeled 'Res block' are standard and bottleneck resnet blocks, respectively, as detailed in [14]. Each fully connected layer is preceded by a dropout layer with probability 0.5, and a softmax non-linearity is used as a final classification layer. The rationale for having a deep low resolution path is to further increase the receptive field and allow for complex higher level features to be learned i.e where an organ is positioned in relation to other structures. To minimize the memory footprint, the high definition path is chosen to be slightly shorter than the low resolution path. This seems reasonable as this path should learn texture information which is likely to require fewer layers and non-linearites. Further details on training hyper-parameters are given in section 3.1.

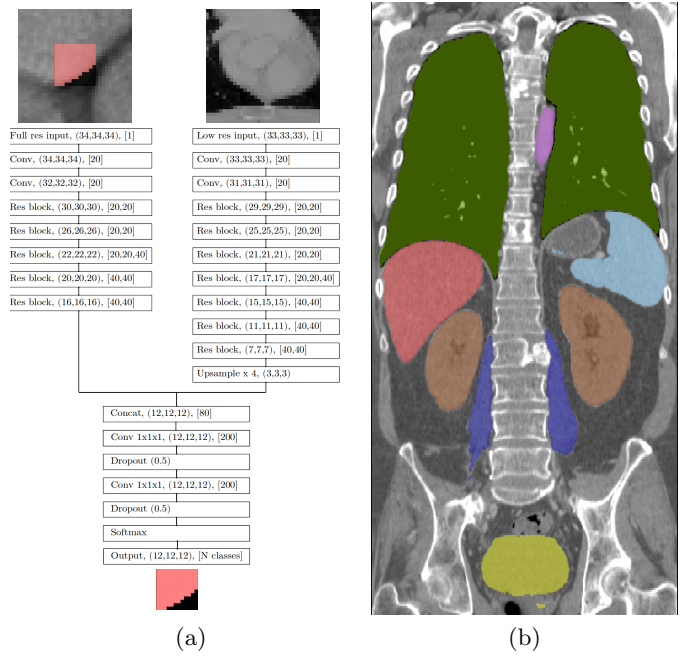

Fig. 1: (a) The proposed dual path CNN architecture. (b) Coronal slice of a CT scan with overlaid segmentation output, described in section 3.3. The organs visible in this slice are: lungs (green), liver (red), spleen (light blue), psoas major muscle (dark blue), kidneys (brown) and bladder (yellow).

## 2.2 Adaptive sampling strategy

The problem of class imbalance as described in [1] can be dealt with by choosing small patch sizes and evenly sampling from each class [1], and through weighted loss functions [3, 15, 16]. Both of these methods either load the whole image into GPU memory, which is not feasible for large images, or select a small subset of patches, which can lead to inefficient training on sparse datasets. To overcome both of these issues we propose the simple sampling Algorithm 1.

---

**Algorithm 1** isample: adaptive sampling algorithm

---

 Initialize error maps for every image in the training data: $\boldsymbol{E}_i(x) = 1$.
 **while** CNN training **do**
  **while** training for 1 epoch **do**
   **while** filling batch with patches **do**
    Pick an image $\boldsymbol{I}_j$ from the training set $\boldsymbol{I}^*$.
    Pick a class $k$ from the corresponding label map $\boldsymbol{L}_j$.
    Pick a patch in image $\boldsymbol{I}_j$, centered at location $\boldsymbol{c}$, where $\boldsymbol{L}_j(\boldsymbol{c}) = k$.
    Accept patch into batch if $\boldsymbol{E}_i(\boldsymbol{c}) > \mathcal{U}(0,1) - \epsilon$.
   **end while**
   Back-propagate loss of batch and update the current CNN weights: $\boldsymbol{w}$.
  **end while**

  Select a subset of images, $\boldsymbol{I}^*$, and label maps, $\boldsymbol{L}^*$, from the training set:
  **for** $[\boldsymbol{I}_k, \boldsymbol{L}_k] \in [\boldsymbol{I}^*, \boldsymbol{L}^*]$ **do**
   Update error maps: $\boldsymbol{E}_k(\boldsymbol{x}) = 1 - \text{CNN}(\boldsymbol{w}, \boldsymbol{I}_k(\boldsymbol{x}))_{\boldsymbol{L}_k(\boldsymbol{x})}$
  **end for**
 **end while**

---

In Algorithm 1, $\mathcal{U}(0,1)$ is a random number drawn from the uniform distribution and $\boldsymbol{E}_i$ refers to the error map of the $i^{th}$ training image. Error maps can easily be calculated, either after each epoch or concurrently to the training process, as

$$\boldsymbol{E}_k(\boldsymbol{x}) = 1 - \text{CNN}(\boldsymbol{w}, \boldsymbol{I}_k(\boldsymbol{x}))_{\boldsymbol{L}_k(\boldsymbol{x})}, \qquad (1)$$

where $\text{CNN}(w, \boldsymbol{I}_k(\boldsymbol{x}))_{\boldsymbol{L}_k(\boldsymbol{x})}$ is a map of the CNN predictions over the full training image $\boldsymbol{I}_k$, evaluated using the most current weights, $\boldsymbol{w}$, and outputting the probability of the true class label $\boldsymbol{L}_k(\boldsymbol{x})$, at position $\boldsymbol{x}$. Examples of error maps produced throughout training are show in Figure 3. The additional parameter $\epsilon$ controls the strength of the isample scheme. Setting $\epsilon = 0$, corresponds to choosing patches based entirely on the amount of error that they currently produce by the network. When $\epsilon = 1$ the condition $\boldsymbol{E}_i(\boldsymbol{c}) > \mathcal{U}(0,1) - \epsilon$ is always satisfied and we are left with a standard sampling scheme that accepts every chosen patch. For all results shown in this paper we have chosen $\epsilon = 0.01$, since we are interested in using the isample scheme to full effect. Detailed investigations into how to best set this parameter for different datasets with varying amounts of sparsity is left for future work. The subset of images, $\boldsymbol{I}^*$, and label maps, $\boldsymbol{L}^*$, of the training set may be chosen in line with how quickly to introduce the isample scheme during training and the amount of computational resources available. In our experiments we had access to four GPUs, three were used to train the CNN continuously and one GPU was used in parallel to continuously perform full predictions of the validation dataset and the training dataset. From this, full dice scores of the validation dataset and full error maps of the training dataset could be calculated. A future extension of this work could log the loss of individual batches during training, and use this information to get an estimate of where the network is having difficulty. This modification would avoid the need for having to allocate additional resources for segmenting the training data, which instead could be used to speed up training. However for practical reasons, and the increasing availability of large multi-GPU cards, we have found this not to be an issue. Also, having access to dice scores calculated over the full images throughout training has been helpful in development since these true dice scores provide more meaningful information than dice scores calculated from individual batches which are biased by the sampling scheme and the size of patches. The error maps produced can also be useful for debugging purposes during development.

## 3 Results

We trained, validated and tested the automatic segmentation method on contrast enhanced CT scans from the VISCERAL Anatomy 3 dataset, made up of 20 training scans, and 10 unseen testing scans (currently not available to download) [17]. The scans are form a heterogeneous dataset with various topological changes between patients, and manual segmentations are available for a number of different anatomical structures. We randomly split the training set into 16 scans for training (80%) and 4 scans for validation (20%), we also present results of our online submission on the unseen test dataset. For illustrative purposes, the first experiment, in section 3.2, focuses on segmenting only the kidneys from full body CT scans. In section 3.3 we present results on simultaneously segmenting multiple organs from the CT data.

### 3.1 CNN training setup

During training we perform data augmentation by re-sampling the 3D patches to a $[1mm, 1mm, 1.5mm] + \mathcal{U}(-0.1, 0.1)$ resolution. We also rotate each patch by $[\mathcal{U}(-10, 10), \mathcal{U}(-4, 4), \mathcal{U}(-4, 4)]$ degrees. We set voxels with values greater than 1000 to 1000, and values less than $-1000$ to $-1000$, and divide all values by a constant factor of 218 (the standard deviation of the dataset). We use Glorot initializations [18] on all convolution layers. For batchnorm layers we use the initializations technique described in [19]. We impose $L_2$ weight decay of size 0.0001, on all convolutional layers except on the last fully convolutional layer before the final softmax non-linearity. Using techniques described in [19] we make use of large batch sizes and large learning rates. We use SGD with Nestrov momentum set at 0.8 [20], the initial learning rate is 0.001, and each batch contains 12 patches, sampled from one randomly selected scans in the training set. We run each epoch for 100 batches. We also employ a learning rate

6       Lorenz Berger et al.

warm up schedule as described in [19] for the first 5 epochs. We use a standard cross-entropy loss function.

### 3.2 Segmenting the kidneys from full body CT scans

In this experiment we use labels for kidneys to train the CNN, resulting in a simple two class, foreground (kidneys) and background (everything else), segmentation problem. Figure 2 shows curves of the training loss (2a) and mean validation dice score (2b) for segmented kidneys throughout training, averaged over three separate runs. The blue curve represents training runs where patches are sampled randomly but evenly from background and kidney foreground, the red curve represents training runs where patches are sampled using the proposed sampling algorithm 1. Because isample adaptively selects more difficult patches as training progresses, the loss is higher, as seen in Figure 2a. In Figure 2b, the sampler achieves faster generalization, and our current results indicate that the final generalization of the CNN trained with the proposed sampling scheme is slightly improved for this sparse segmentation setup, where the kidneys only make up $\sim 0.3\%$ of the voxels within the whole scan.

Table 4a shows the average dice scores achieved by the Dual CNN, with and without isample, throughout training. When using the isample scheme the CNN is able to achieve a dice score of 0.855 after only $5k$ training iterations. This is close to the end of training performance, a dice score of 0.899 after 40k, of the Dual CNN without isample in use. Fast training of CNNs using isample can be useful for debugging and evaluating changes in neural network architectures, as more experiments can be run using the same computational resources, that quickly estimate the end performance of the network.

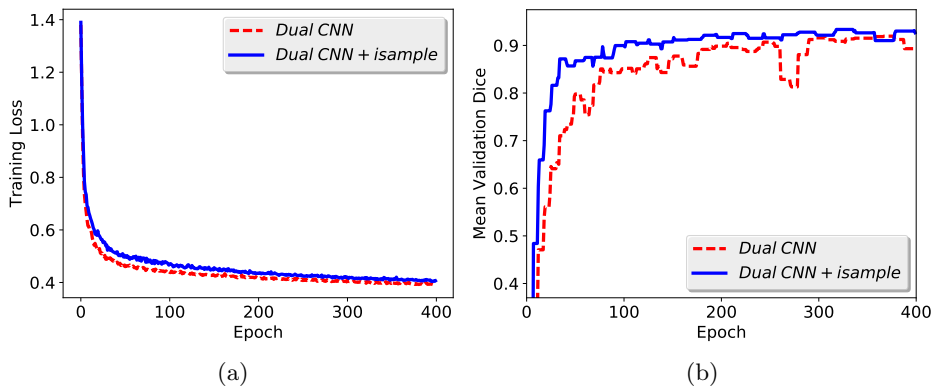

Fig. 2: (a) Training loss (a) and mean validation dice scores (b), averaged over 3 runs.

In Figure 3 we show coronal slices of a training error volume $\boldsymbol{E}_k(\boldsymbol{x})$, calculated by the algorithm. As seen in Figure 3c, initially there is significant produced by the CNN prediction at epoch 16, for example misclassifying the aorta (part of the background class) because it has similar intensity values to the kidneys. After more training, at epoch 50, Figure 3d shows that the error is much lower. The CNN has now learned that the aorta is part of the background class. However more subtle regions such as the collecting system and large vessels within the kidney (see small hole in the true segmentation of the left kidney in Figure 3b) still produce high errors, and further focused training is required to optimize the weights until they are correctly classified. There also remains a high error around the border of the kidneys, which will result in the sampling process selecting more patches from the border region, and thus ends up learning to train the network with a similar loss to the hand-crafted border weighted loss function designed in [3].

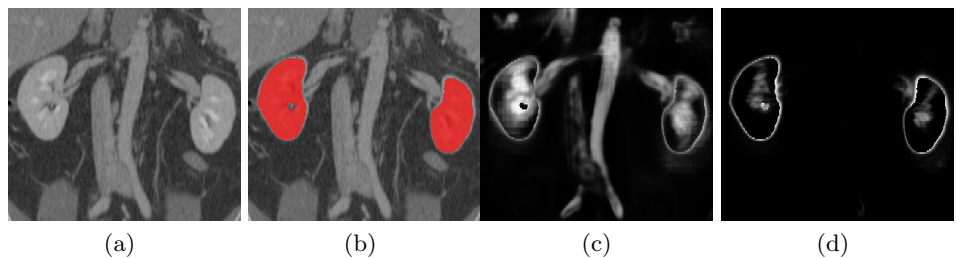

(a)      (b)      (c)      (d)

Fig. 3: Coronal slices: (a) Raw CT scan from the training set (b) Kidney segmentation overlaid onto scan (c) Error map, $\boldsymbol{E}_k(\boldsymbol{x})$, of foreground and background classification on a training scan after 16 epochs. (d) $\boldsymbol{E}_k(\boldsymbol{x})$ after 50 epochs. For the error maps, white corresponds to voxels that are incorrectly classified and black to correctly classified voxels.

| # its | Dual CNN | + isample |
|---|---|---|
| 5k | 0.797 | 0.855 |
| 10k | 0.849 | 0.897 |
| 20k | 0.905 | 0.920 |
| 40k | 0.899 | 0.927 |

(a)

| Method | Kidney Dice scores | |
|---|---|---|
| Dual CNN (validation data) | Left + Right 0.899 (0.058) | |
| Dual CNN + isample (validation data) | Left + Right 0.927 (0.037) | |
| Dual CNN + isample + CRF (test data) | Left 0.954 | Right 0.96 |
| Wang et al [21] (test data) | Left 0.945 (0.027) | Right 0.959 (0.011) |
| Vincent et al [22] (test data) | Left 0.943 (0.015) | Right 0.927 (0.040) |
| Gass et al [23] (test data) | Left 0.913 (0.029) | Right 0.914 (0.027) |

(b)

Fig. 4: (a) Mean dice scores and standard deviations at different number of iterations, throughout training. (b) Dice scores and standard deviations, where available, for different methods automatically segmenting kidneys on the VISCERAL CT enhanced dataset.

Table 4b shows dice scores for segmenting both kidneys using different methods. The proposed method with isample performs significantly better than without. We also submitted our method, with the addition of a CRF [24] as a post-processing step, to segment the test dataset, and achieved the top score for segmenting the left and right kidneys. Inference on a full size CT scans takes ∼ 65 seconds using four Tesla K50 GPU cards, each with 4GB of RAM.

### 3.3  Multi-organ segmentation

We now extended the previously described algorithm to include a multi-class classification output and trained the model on the main organs available on the VISCERAL CT-enhanced dataset. We post-processes the output segmentation maps (maximum class probability at each voxel), by applying a filter that only retains the largest connected binary object within the segmentation, thus removing small objects. The segmentation output of one of the the validation scans is shown in Figure 1b. The results of our proposed method and other state-of-the-art methods, also summarized in [17], are given in Table 1.

| Method | Aorta | Lung | Kidney | PMajor | Liver | Abdom | Spleen | Sternum | Trachea | Bladder |
|---|---|---|---|---|---|---|---|---|---|---|
| Dual CNN + isample (val) | 0.893 | 0.980 | 0.938 | 0.824 | 0.941 | 0.769 | 0.951 | 0.900 | 0.926 | 0.912 |
| Dual CNN (val) | 0.843 | 0.985 | 0.934 | 0.779 | 0.927 | 0.755 | 0.946 | 0.904 | 0.926 | 0.918 |
| Ga et al [23] | 0.785 | 0.963 | 0.914 | 0.813 | 0.908 | - | 0.781 | 0.635 | 0.847 | 0.683 |
| Jimenez et al [25] | 0.762 | 0.961 | 0.899 | 0.797 | 0.887 | 0.463 | 0.730 | 0.721 | 0.855 | 0.679 |
| Kéchichian et al [26] | 0.681 | 0.966 | 0.912 | 0.802 | 0.933 | 0.538 | 0.895 | 0.713 | 0.824 | 0.823 |
| Vincent et al [22] | 0.838 | 0.972 | 0.935 | 0.869 | 0.942 | - | - | - | - | - |
| Inter-annotator agreement | 0.859 | 0.973 | 0.917 | 0.823 | 0.965 | 0.673 | 0.934 | 0.810 | 0.877 | 0.857 |

Table 1: Dice scores for different automatic multi-organ segmentation methods and inter-annotator agreement results [17] on the VISCERAL dataset.

We note that because the cloud-based evaluation service [17] containing the test data was closed at the time of running these experiments, we were not able to evaluate our method on the test data, thus making direct comparisons to previous methods difficult. As previously mentioned, we trained our method on 80% of the data (16 scans) and validated it on the remaining 20% (4 scans). From having evaluated the kidney only CNN on the test data, we found that the testing dataset gave better dice scores than the validation set. We are therefore confident that our results in Table 1 are representative. As soon as the test data set is made available we will update our results. Also for organs 'Lung', 'Kidney', 'PMajor' (psoas major) and 'Abdom' (rectus abdominis) we give mean dice scores of both the left and right organs. For this experiment we modified the training schedule such that the initial learning rate is 0.05, and each batch contains 24 patches, sampled from 2 randomly selected scans from the training set. We run each epoch for 200 batches, and halve the learning rate after every 25 epochs. The Dual CNN without using the isample scheme (average organ dice 0.8917) slightly underperformed compared to when using the isample (average organ dice 0.9034). However this difference is far less notable than during previous experiments, shown in Table 4b. We hypothesize this is because the background class in the multi-organ segmentation problem is split into background and other organs such as the Lung, Liver etc, thus making the dataset, especially the background class, easier to sample from. The potential benefit of using the isample method is therefore problem dependent.

## 4  Conclusion

We proposed and evaluated a sampling scheme to deal with very large images such as 3D CT scans. As shown in section 3 the sampler enables fast training, and our results indicate that the final generalization performance can be improved. This is inline with previous research that shows the positive effect of curriculum learning on optimization and end performance of machine learning systems [9, 8]. Our experimental results suggests our algorithm gives new state of the art performance for the aorta, lung, kidney, rectus abdominis, spleen, sternum, trachea and bladder, on the VISCERAL anatomy benchmark, and improves upon human inter-annotator agreement scores on the following organs: aorta, lung, kidney, psoas major, rectus abdominis, spleen, sternum, trachea and bladder. These encouraging results pave the way for using CNNs for robust automatic segmentation within clinical practice, such as surgical planning.

10      Lorenz Berger et al.